\documentclass[letterpaper]{article} 
\usepackage{aaai25}  
\usepackage{times}  
\usepackage{helvet}  
\usepackage{courier}  
\usepackage[hyphens]{url}  
\usepackage{graphicx} 
\usepackage{natbib}  
\usepackage{caption} 
\usepackage{algorithm}
\usepackage{algorithmic}
\graphicspath{{figures/}}
\usepackage{amsmath}
\usepackage{epsfig}
\usepackage{amssymb}
\usepackage{pifont}      
\usepackage{booktabs}

\usepackage{newfloat}
\usepackage{listings}
\usepackage{epsfig}
\usepackage{amsmath}
\usepackage{multirow}
\usepackage{multicol}
\usepackage{booktabs}
\usepackage{xcolor}
\DeclareCaptionStyle{ruled}{labelfont=normalfont,labelsep=colon,strut=off} 
\DeclareCaptionStyle{ruled}{labelfont=normalfont,labelsep=colon,strut=off} 
\floatstyle{ruled}
\newfloat{listing}{tb}{lst}{}
\floatname{listing}{Listing}
\title{Multi-task Image Restoration Guided by Robust DINO Features}

\author {
    Xin Lin$^{1,2}$ \quad Jingtong Yue$^{1,2}$ \quad Kelvin C.K. Chan $^{3}$ \quad Lu Qi $^{2}$\\
    Chao Ren$^{1}$ \quad Jinshan Pan $^{4}$\quad Ming-Hsuan Yang $^{2,3}$
}
\affiliations{
    $^{1}$Sichuan University\\
    $^{2}$UC Merced\\
    $^{3}$Google Research\\
    $^{4}$Nanjing University of Science and Technology
}

\usepackage{bibentry}

\begin{document}

\maketitle

\begin{abstract}
Multi-task image restoration has gained significant interest due to its inherent versatility and efficiency compared to its single-task counterpart. However, performance decline is observed with an increase in the number of tasks, primarily attributed to the restoration model's challenge in handling different tasks with distinct natures at the same time. Thus, a perspective emerged aiming to explore the degradation-insensitive semantic commonalities among different degradation tasks. In this paper, we observe that the features of DINOv2 can effectively model semantic information and are independent of degradation factors. Motivated by this observation, we propose \mbox{\textbf{DINO-IR}}, a multi-task image restoration approach leveraging robust features extracted from DINOv2 to solve multi-task image restoration simultaneously. We first propose a pixel-semantic fusion (PSF) module to dynamically fuse DINOV2's shallow features containing pixel-level information and deep features containing degradation-independent semantic information.
To guide the restoration model with the features of DINOv2, we develop a DINO-Restore adaption and fusion module to adjust the channel of fused features from PSF and then integrate them with the features from the restoration model. By formulating these modules into a unified deep model, we propose a DINO perception contrastive loss to constrain the model training. 
Extensive experimental results demonstrate that our DINO-IR performs favorably against existing multi-task image restoration approaches in various tasks by a large margin.
The source codes and trained models will be made available.
\end{abstract}

%

\section{Introduction}
\label{sec:intro}

Image restoration aims to recover a clean image from its degraded counterpart. 
Significant progress has been made in each restoration task (denoising \cite{deamnet, scaoednet, scpgabnet, pyz, pyzcsvt}, deblurring \cite{deblurring1, deblurring2, deblurring3, deblurring4}, and deraining \cite{llrr, did, didmdn, sirr, lpnet}, among others \cite{ecai, sr1, sr2, sr3, sr4, sr5, sr6} with the development of deep learning. 
%
However, when solving a different degradation images, they have to redesign or retrain existing deep models to achieve satisfactory results. 
This inevitably increases computational overhead and storage requirements when handling multiple degradations.

\begin{figure}[t]
  \begin{minipage}[t]{0.45\textwidth}
    \centering
    \includegraphics[width=\textwidth]{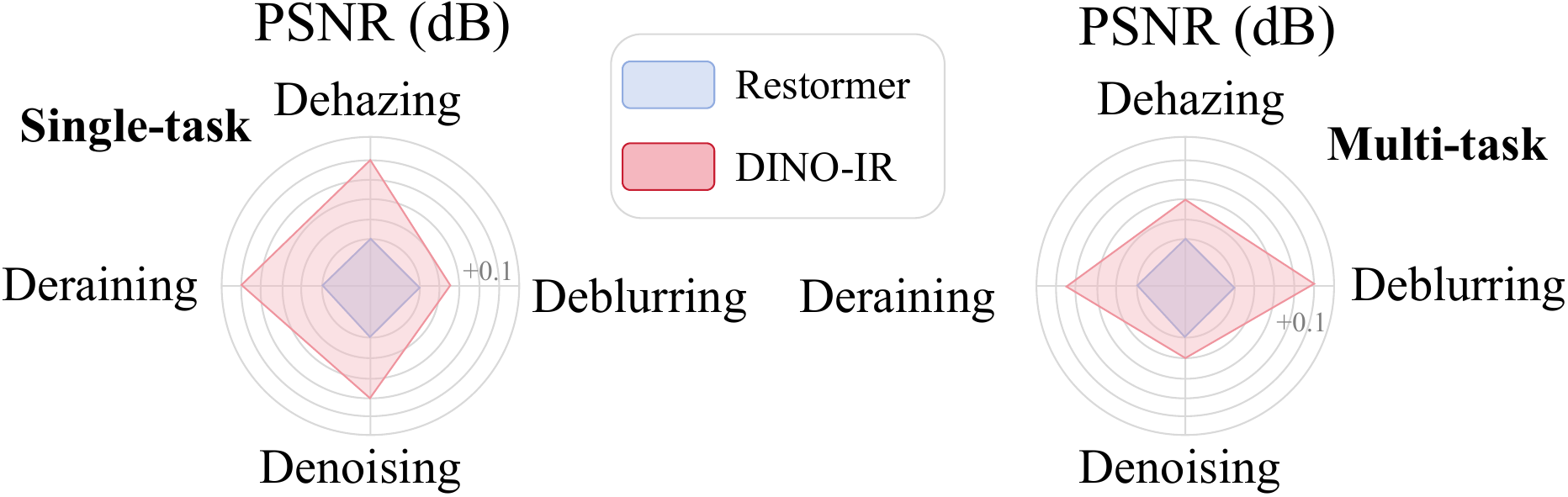}
    (a) Comparison of Restormer and DINO-IR.
  \end{minipage}\hfill
  \begin{minipage}[t]{0.45\textwidth}
    \centering
    \includegraphics[width=\textwidth]{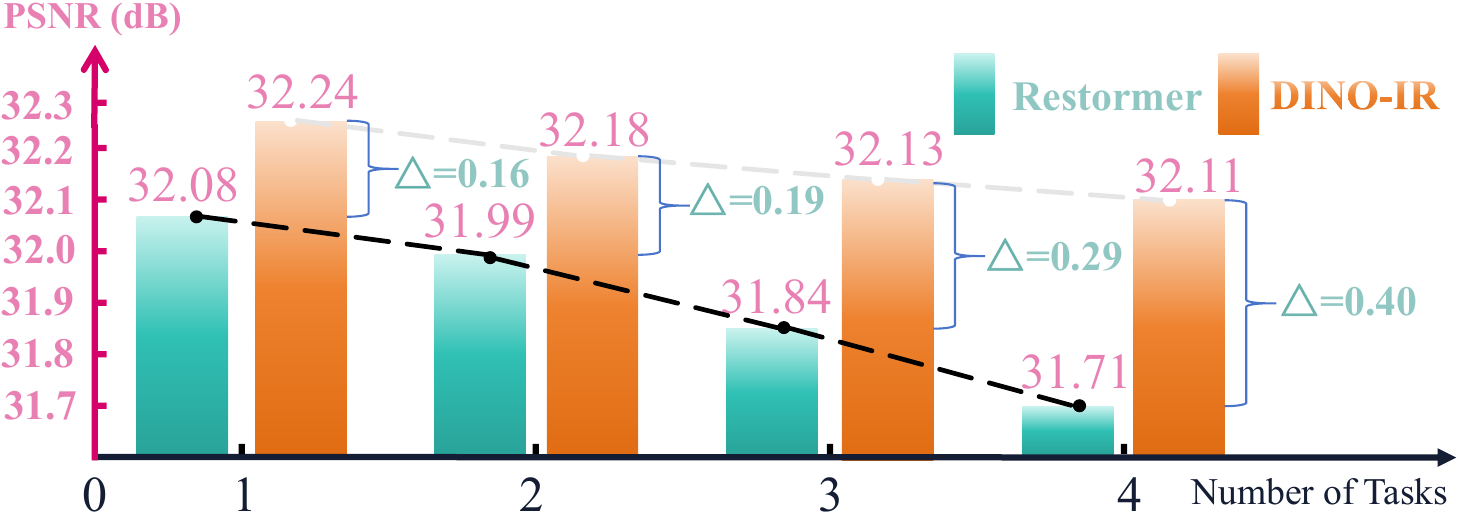}
    (b) Comparison of Restormer and DINO-IR.

  \end{minipage}
  \caption{(a) Results from both single-task and multi-task image restoration for Restormer \cite{restormer} and our DINO-IR, with ours demonstrating superior performance in these tasks. (b) Comparison of the performance of our method and Restormer in the deblurring task as the number of restoration tasks increased. Both DINO-IR and Restormer experience a decline in performance. However, due to the stable features provided by DINOv2, our method exhibits enhanced stability, mitigating the extent of the performance decline and ultimately surpassing the baseline.}
  \label{dongji}
\end{figure}

%
Multi-task image restoration \cite{TKMANet, airnet, idrnet, acmmm, wgws, daclip, promptir} has emerged as a promising approach to solve the aforementioned challenges. 
These methods train one model from scratch to handle multiple degradations, requiring the model to discriminate different types of degradation and provide corresponding restoration. 
Although this strategy improves cost-effectiveness and versatility, the model must adapt to various restoration tasks, each with unique characteristics. 
For example, deblurring aims at enhancing high-frequency details, whereas denoising targets to remove high-frequency noise.
Thus, it is challenging to learn representations of different conflicting tasks at the same time.
As shown in Fig. \ref{dongji}(b), the performance of existing methods deteriorates significantly when the number of tasks increases. 
Therefore, it is important to explore degradation-agnostic representations for robust multi-task image restoration.

DINOv2 \cite{dinov2}, a large-scale pre-trained model that can provide universal representations, giving powerful priori for the downstream tasks \cite{junyi, dino1, foundation1}. 
We find that the deep layers of DINOv2 provide robust semantic representations insensitive to degradations while retaining high-frequency contour details. Based on this observation, we exploit DINOv2 to learn the degradation-independent features of different degraded images in multi-task image restoration. 
Specifically, we analyze the robust, degradation-agnostic semantic features from DINOv2 in multi-task restoration and the effect of different models and feature layers of DINOv2 on a range of degraded images.
In conjunction with its shallow features, which capture low-level image details, \mbox{DINOv2} emerges as a promising candidate for multi-task image restoration.

To better explore features from \mbox{DINOv2}, we develop an effective \textbf{DINO-IR} including 1) Pixel-semantic fusion (PSF) module, 2) DINO-Restore (D-R) adaption and fusion module, and 3) DINO perception contrastive (DPC) loss for multi-task image restoration. We show that the proposed \textbf{DINO-IR} enhances model performance while reducing its sensitivity to degradations across multiple tasks. 
The main contributions of our work are:
\begin{itemize}
        \item We develop an effective PSF module to dynamically fuse DINOV2’s shallow features containing pixel-level information and deep features containing semantic, degradation-independent information. 
        \item We develop a D-R adaption and fusion module that integrates the DINO features by the PSF module into the image restoration model for better image restoration with different degradations.
        \item Leveraging the ability of shallow features from DINOv2 to characterize low-level image details, we present DINO perceptual contrast learning to constrain the training process and show that the proposed method achieves favorable performance against state-of-the-art ones.
\end{itemize}

\section{Related Work}
\label{Related Work}

{\flushleft \textbf{Multi-Task Restoration.}} 
Multi-task restoration is focused on empowering a simple model to address images affected by multiple types of degradation.
All in One \cite{allinone} tackles various bad weather degradations with a multi-encoder and single-decoder framework.
On the other hand, IPT \cite{duotou} employs a multi-head and multi-tail architecture based on the transformer to handle a variety of degradation scenarios. 
After that, TransWeather \cite{transweather} utilizes weather-type queries to address diverse degradation issues within a single encoder-decoder transformer framework. 
Recently, AirNet \cite{airnet} introduces a prior-free network with contrastive learning that does not distinguish between corruption types and ratios, and IDRNet \cite{idrnet} presents a novel perspective that explores degradation through an ingredient-oriented approach, ultimately improving the model's scalability. 
A learning network based on degradation classification is introduced in AMIRNet \cite{acmmm} by
utilizing its robust classification capabilities to guide the restoration process effectively. 
WGWS \cite{wgws} dynamically incorporates parameters tailored to distinct weather types during the latter phase, thereby addressing a spectrum of weather conditions. 
Most recently, PromptIR \cite{promptir} presents a prompt-based learning approach that effectively restores images affected by various types and levels of degradation. 
DA-CLIP \cite{daclip} integrates a large-scale pre-trained visual-language model CLIP \cite{clip} in image restoration tasks. 
It concentrates on the identification of degradation categories through text information. However, the above methods lack the exploration of robust feature guidance for various restoration tasks.
\begin{figure*}[t]
\centering
\includegraphics[width=1\linewidth]{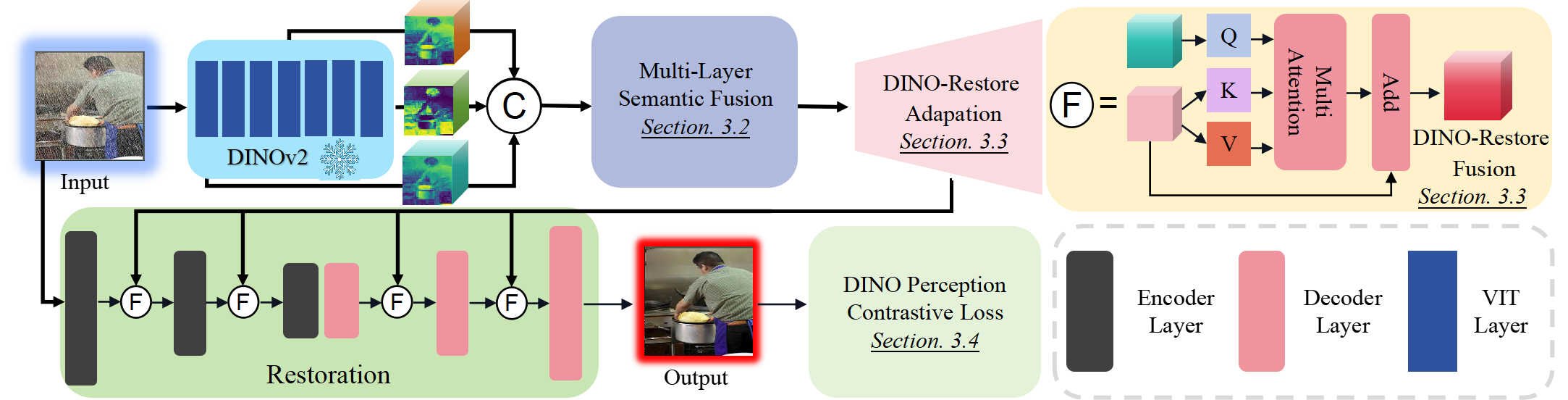}
\caption{\label{kuangjia} 
DINO-IR framework comprising the following components: 1. Pixel-semantic fusion (\textbf{PSF}) module and DINO-Restore (\textbf{D-R}) adaption and fusion module: They facilitate better fusion of shallow, medium, and deep DINOv2 features to guide restoration. 2. \textbf{Restoration}: This component takes low-quality images as input and utilizes a restoration network to return them to a clean version. 3. DINO perception contrastive (\textbf{DPC Loss}): This loss function enhances performance through DINOv2 feature contrastive learning.}
\end{figure*}

{\flushleft \textbf{Pre-training of Vision Models.}}
With the advances of deep learning, numerous visual models such as Beit \cite{beit}, CLIP \cite{clip}, BLIP \cite{blip}, MAE \cite{mae}, SimMIM \cite{mim}, DINO \cite{dino}, DINOv2 \cite{dinov2} emerge. 
These models are typically pre-trained on ImageNet and fine-tune for downstream tasks, consistently delivering outstanding performance in diverse domains.
CLIP~\cite{clip} is a language-supervised pre-training approach that effectively integrates images and text, achieving remarkable zero-shot transfer capabilities. 
MAE \cite{mae} and SimMIM \cite{mim} employ image pixel reconstruction to imbue pre-trained networks that adapt to a broad spectrum of downstream tasks. 
DINO-ViT \cite{dino}, a self-supervised ViT model training method, demonstrates its effectiveness in multiple tasks, including image retrieval and object segmentation. 
Recently, DINOv2 \cite{dinov2} enhances the model scale of DINO and augments it with a larger volume of data. 
Amir et al. and Zhang et al. \cite{junyi, dino1} verify that using the features of DINO and DINOv2 provides robustness and strong generalization to guide downstream tasks, such as sparse correspondence, dense correspondence, and instance swapping. These related works inspired us to explore DINOv2 in low-level image restoration tasks.

\section{Propsed Method}

In this section, we first analyze the property of the features from DINOv2 on different image degradations and then present the proposed \textbf{DINO-IR} to utilize the features from DINOv2 for multi-task image restoration.

\subsection{Motivation}

\begin{figure}[t]
  \begin{minipage}[t]{0.49\textwidth}
    \centering
    \includegraphics[width=\textwidth]{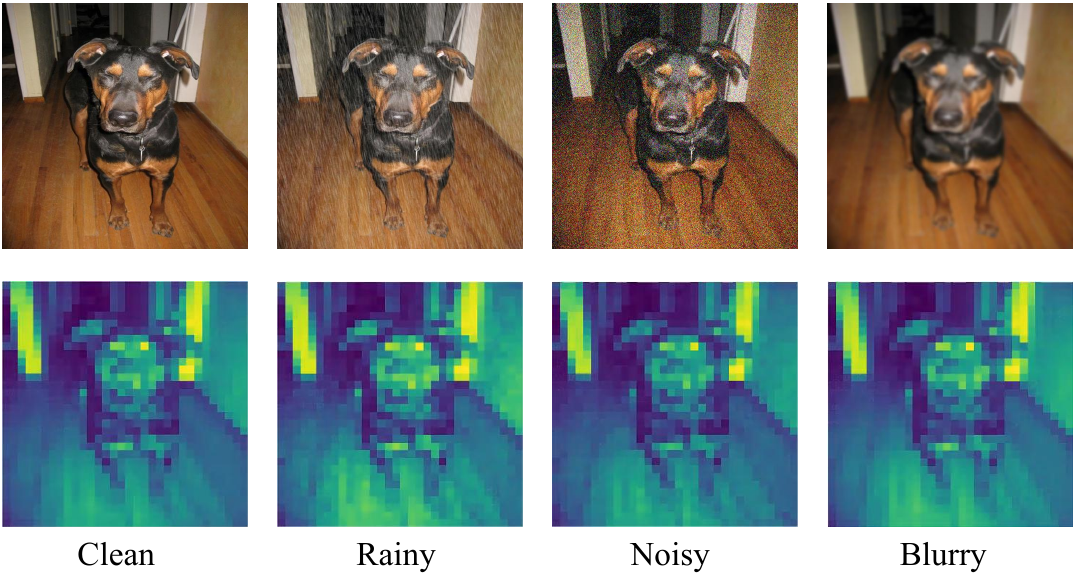}
    (a) Visualization of DINOv2 features for different degradation images.
  \end{minipage}\hfill
  \begin{minipage}[t]{0.49\textwidth}
    \centering
    \includegraphics[width=\textwidth]{./image/zhuzhuangtu_2}
    (b) The deviation of PSNR for image, \(f_{\mathrm{IMAGE}}\) and \(f_{\mathrm{DINO}}\).
  \end{minipage}
  \caption{(a) We obtain rainy, noisy, and blurry images by degrading a clean version in three ways. Following this, we extract features from these images by the deep layer of DINOv2 and visualize them by feature projection as used in \cite{dino1}. (b) We compare the variations in the deviation of PSNR for Image, \(f_{\mathrm{IMAGE}}\), \(f_{\mathrm{DINO}}\) as the noise level increases.} 
  \label{duojiangzhi}
\end{figure}

We analyze the robustness of DINOv2 features against various degraded images through qualitative and quantitative comparison. 
We corrupt a clean image by different types of degradation (rain, noise, and blur) and pass it to the DINOv2 backbone. 
We then apply principal components analysis as referenced in \cite{dino1, junyi} to the features. 
As depicted in Fig. \ref{duojiangzhi}(a), the clean image's DINOv2 feature effectively preserves the structure contours information. 
Meanwhile, the DINOv2 features obtained from the degraded images are highly similar to those from the clean image. 
This indicates that DINOv2 features are stable across different types of degradation while retaining essential semantic information.

For illustration, we compute DINOv2 features \(f_{\mathrm{DINO}}\) and degraded image features (generated by patch embedding) \(f_{\mathrm{IMAGE}}\) under varying noise conditions, and compute their PSNR to the clean counterpart.
We assess the stability of DINO features using the difference between the PSNR at a specific noise level and the mean PSNR across all noise levels. 
Additionally, we compute the variance of PSNR across different noise levels. 
From Fig.~\ref{duojiangzhi}(b), \(f_{\mathrm{DINO}}\) (blue bar) possesses a significantly smaller deviation compared to the degraded image (green bar) and \(f_{\mathrm{IMAGE}}\) (orange bar) across most evaluated noise distributions. Furthermore, the variance for \(f_{\mathrm{DINO}}\) is 0.61, markedly lower than the 10.47 of \(f_{\mathrm{IMAGE}}\) and 36.85 of the degraded image, showing that DINO features are more robust against degradations. This observation suggests these features have semantic and robust qualities and their rich contour features are advantageous in restoring prominent edge information. Additional analyses can be found in the supplementary material.

\begin{figure}[t]
\centering
\includegraphics[width=1\linewidth]{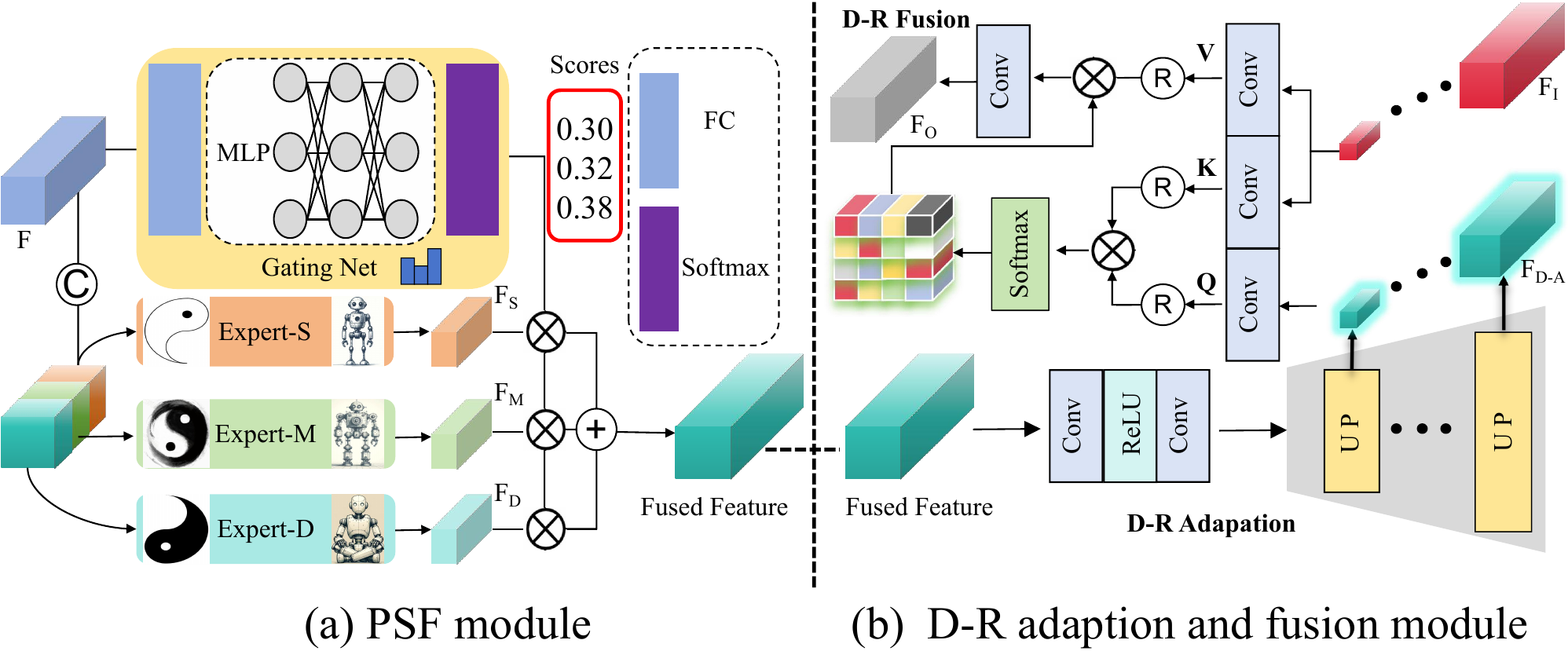}
\caption{\label{modules} (a) The architecture of the Pixel-semantic fusion (PSF) module. PSF is employed to fuse the shallow pixel-level and deep semantic features of DINOv2. (b) The architecture of DINO-Restore (D-R) adaption and fusion module. Its task is to adjust the size of DINOv2 features to fit the restoration network and then fuse with the feature from the restoration network by a self-attention-based method.}
\end{figure}


%

\subsection{Pixel-Semantic Fusion (\textbf{PSF}) Module}
\label{section_msf}
DINOv2 is hierarchical, with each layer presenting unique characteristics essential for restoration. As is widely recognized, shallow layers possess pixel-level information, while deep layers hold high-level semantic attributes. This understanding informs our approach to use both to guide restoration. It is imperative to selectively combine and refine these hierarchical features, giving more weight to those layers that greatly influence restoration. For this purpose, we design a Pixel-semantic fusion (PSF) module to exploit shallow, medium, and deep features. 
As shown in Fig. \ref{modules}(a), the PSF module operates on dynamic and conditional inputs, employs specialized experts, and executes collaborative processing, fitting perfectly for the diverse and distinct layers of DINOv2. It consists of a gating network alongside multiple expert networks. The adaptive and learnable gating network uses feedback from DINOv2's various layer features to steer the feature learning process and determine the weights for output features of the expert networks.
Features advantageous for the restoration model are assigned higher weights, while those less beneficial receive lower weights. Specifically, we set three CNN-based experts: $\mathbf{\mathrm{E_{S}}}$, $\mathbf{\mathrm{E_{M}}}$, and $\mathbf{\mathrm{E_{D}}}$, which correspond to the shallow, medium, and deep features from DINOv2. An efficient MLP-based gating network to compute the importance of three types of DINOv2 features and output the scores. Due to the limited space, the specific structure of modules is provided in the supplementary material.

The process is as follows:
\setlength{\abovedisplayskip}{4pt} 
\begin{align}
&\mathbf{F}  = \mathrm{Concat}\left[\mathrm{DINO_S}(\mathbf{I}) \hspace{1mm} \mathrm{DINO_M}(\mathbf{I}) \hspace{1mm} \mathrm{DINO_D}(\mathbf{I})\right], \\
&\mathbf{S_{S,M,D}}  = \mathrm{Gating}\left(\mathbf{F}\right), \\
&\boldsymbol{\mathrm{F_{S,M,D}}} = \boldsymbol{\mathrm{E_{S,M,D}}} (\mathrm{DINO_S}(\mathbf{I}),\mathrm{DINO_M}(\mathbf{I}),\mathrm{DINO_D}(\mathbf{I})), \\
&\boldsymbol{\mathrm{F_{Fused}}} = \boldsymbol{\mathrm{F_S}} \times \boldsymbol{\mathbf{S_S}} + \boldsymbol{\mathrm{F_M}} \times \boldsymbol{\mathbf{S_M}} + \boldsymbol{\mathrm{F_D}} \times \boldsymbol{\mathbf{S_D}},
\end{align}
\noindent where $\mathbf{I}$ is the input image; $\mathrm{DINO_{S,M,D}}$(\textbf{I}) represents DINO's shallow, medium, and deep features, respectively; $\boldsymbol{\mathrm{S_{S,M,D}}}$ and $\boldsymbol{\mathrm{F_{S,M,D}}}$ are the output scores of the gating and expert networks for the shallow, medium, and deep features. 
$\boldsymbol{\mathrm{F_{Fused}}}$ is the fused feature from the $\boldsymbol{\mathrm{S_{S,M,D}}}$ and $\boldsymbol{\mathrm{F_{S,M,D}}}$.

\subsection{D-R Adaption and Fusion Module}
\label{section_ronghe}
As illustrated in Fig. \ref{modules}(b), we implement a simple yet effective adaption and fusion module. Initially, the fused features from the PSF module are fed into the adaption part, where we adjust the channel number and scale to match the restoration model’s configuration. We then employ a self-attention-based fusion approach, inputting DINO features and restoration features of the same scale together where the adapted DINO feature (\textup{F\textsubscript{D-A}}) acts as the query (Q). The restoration feature (\textup{F\textsubscript{I}}) serves as both key (K) and value (V). After the self-attention operation, the residuals are aggregated. This operation is performed through channel-wise multiplication ($O(C^{2}$)). As demonstrated in \cite{restormer}, this technique enhances feature learning and performance while minimizing the demand for computational resources. The procedure is as follows:
\begin{equation}
\mathrm{F_O} = \text{Softmax}\left( \mathrm{W_k} (\mathrm{F_{D-A}}) \times \mathrm{W_q} (\mathrm{F_I}) / \sqrt{\mathrm{C}} \right) \times \mathrm{F_I} + \mathrm{F_I},
\end{equation}

\noindent where $\mathrm{W_k}(\cdot)$ and $\mathrm{W_q}(\cdot)$ are $1 \times 1$ convolution layer, `$\times$' represents matrix multiplication and $\mathrm{C}$ denotes the channel of features. The attention map reflects the inter-relationship between the restoration feature $\mathrm{F_I}$ and the adapted DINO feature $\mathrm{F_D}$. Then, we use the attention map to fabricate restoration features $\mathrm{F_I}$. The $\mathrm{F_O}$ serves as the output of the $b_{th}$ block and the input for the $(b + 1)_{th}$ block within the restoration model.

\subsection{DINO Perception Contrastive Loss} 
\label{dpc}

\begin{figure}[t]
\centering
\includegraphics[width=1\linewidth]{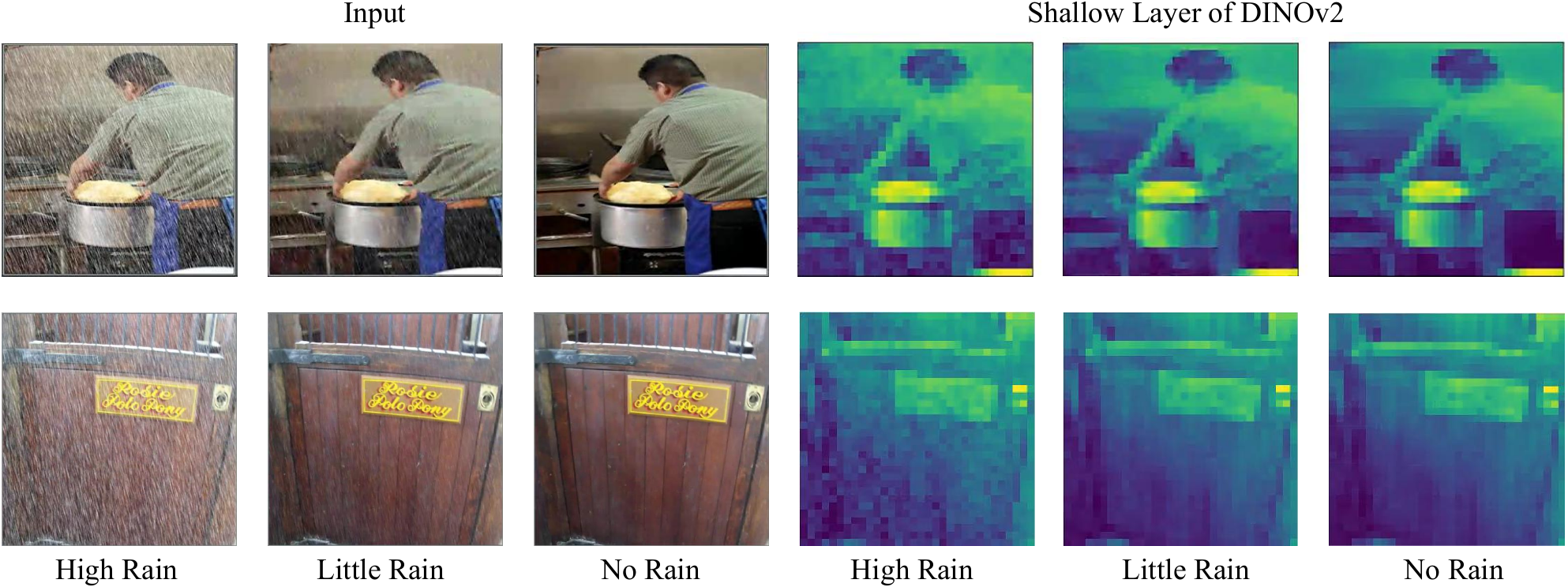}
\caption{\label{duibixuexi} Visualization of the features from the shallow layers of DINOv2 for high rain, light rain, and no rain images.}
\end{figure}

It is known that the features extracted from shallow layers of DINOv2~\cite{dinov2} can discern low- and high-quality images. As shown in Fig. \ref{duibixuexi}, features extracted from the shallow layer can effectively distinguish between high-quality clean and low-quality rainy images. 
When applied to images subjected to initial rain removal processing (``Little Rain''), the features in the DINOv2 representation exhibit a closer alignment with those of clear images.

\begin{table*}[t]
    \centering
    \fontsize{15}{15}\selectfont
    \caption{Multi-task restoration results comparisons. The red color indicates the best results, and the blue color indicates the second-best results. On average, across various tasks, DINO-IR demonstrates improvement of 1 and 0.13 dB over the previously multi-task methods AirNet \cite{airnet} and PromptIR \cite{promptir}.}
    \label{multi}
    \renewcommand{\arraystretch}{1.5}
    \resizebox{\linewidth}{!}{
    \begin{tabular}{cccccccccc}
        \toprule
        \multirow{2}{*}{Method} & \multicolumn{2}{c}{Deraining} & \multicolumn{3}{c}{Denoising} & \multirow{2}{*}{Deblurring} & \multirow{2}{*}{Dehazing} & \multirow{2}{*}{Average} \\
        \cmidrule(lr){2-3} \cmidrule(lr){4-6} 
         & High & Medium & $\sigma = 15$ & $\sigma = 25$ & $\sigma = 50$ & & \\
        \midrule
        DnCNN \cite{dncnn} & 24.57/0.780 & 28.66/0.859& 30.89/0.910& 29.20/0.859& 26.08/0.752& 27.50/0.817& 21.44/0.871 &  
        26.90/0.835\\
        MPRNet \cite{mprnet} &29.77/0.879 & 31.86/0.915& 33.65/0.936& 31.07/0.892& 27.77/0.796& 31.56/0.895& 29.01/0.974 & 30.67/0.898\\
        NAFNet \cite{nafnet} & 30.15/0.887& 32.11/0.920& 33.66/0.936& 31.09/0.894& 27.86/\underline{0.803}& 31.63/0.900& 28.80/0.970& 30.75/0.901\\
        Restormer \cite{restormer} & 30.10/0.887& 32.15/0.920& 33.78/0.936& 31.18/0.892& 27.96/0.800& 31.71/0.899& \underline{29.67}/\underline{0.975} & 30.94/0.901\\
        AirNet \cite{airnet} & 29.30/0.875 & 31.62/0.915 & 33.68/0.937 & 31.07/0.894& 27.67/0.791& 31.16/0.887& 26.45/0.957& 30.13/0.893\\
        WGWS \cite{wgws} & 29.38/0.879 &  31.88/0.917 &  33.47/0.935 & 30.94/0.887 & 27.71/0.792 & 31.56/0.896 & 27.97/0.964 & 30.42/0.896\\
        DA-CLIP \cite{daclip} & 30.17/0.887 & 31.92/0.917 & 31.27/0.913 & 30.27/0.876 & 26.82/0.778 & 31.77/0.900 & 28.98/0.974 & 30.17/0.892\\
        PromptIR \cite{promptir} &\underline{30.31}/\underline{0.889} & \underline{32.23}/\underline{0.921}& \underline{33.82}/\underline{0.937} & \underline{31.21}/\underline{0.895} & \underline{27.97}/0.801 & \underline{31.94}/\underline{0.901} & 29.52/0.974 & \underline{31.00}/\underline{0.902}\\
        \textbf{DINO-IR (Ours)} & \textbf{30.44}/\textbf{0.892}& \textbf{32.37}/\textbf{0.923}& \textbf{33.85}/\textbf{0.938} & \textbf{31.26}/\textbf{0.895} & \textbf{28.04}/\textbf{0.807} & \textbf{32.11}/\textbf{0.903}& \textbf{29.86}/\textbf{0.976} & \textbf{31.13}/\textbf{0.904}\\
        \bottomrule
    \end{tabular}}
\end{table*}

In addition, the different layers of DINOv2 exhibit hierarchical variations, which allow the model to focus on different aspects due to the diverse characteristics of these variations. 
As used in \cite{duibiquwu1, duibidehazing} methods, the latent feature space of different layers enables better extraction of information from the image, resulting in better image quality discrimination. 

These results suggest that these features can be harnessed to assess model quality and serve as a loss function to improve visual performance and overall capabilities. 
We propose a contrastive learning approach: 
\begin{equation}
\label{ssim}
L_{DINO} = L(v, v_{+}, v_{-})=\sum_{i=1}^n w_{i}\dfrac{D(\Psi_{i}(v), \Psi_{i}(v_{+}))}{D(\Psi_{i}(v), \Psi_{i}(v_{-}))},
\end{equation}

\noindent where $\Psi_{i}$ ($i = 1,2,...,n$)  extracts the $i_{th}$ hidden features from the fixed pre-trained DINOv2; $D(x, y)$ is the $L_{1}$ distance between $x$ and $y$ and $v$ denotes the restored output image. 
The positive samples $v_{+}$ are clean target images, while the negative samples $v_{-}$ are low-quality input images. 
The objective is for the features extracted from the restored image `output' by DINOv2 to be close to those from the clean target image and far from those extracted from the low-quality input image.
Thus, the overall loss can be expressed as follows:
\begin{equation}
\label{L}
L= L_{1} + \lambda L_{DINO},
\end{equation} 
where $ \lambda$ denotes a weight parameter. 

\section{Experimental Results}
\label{Experiments}

\begin{table*}[ht]
    \centering
    \fontsize{7}{7}\selectfont
    \caption{Denoising comparisons in the single-task setting on BSD68 \cite{bsd68} and Urban100 \cite{urban100} datasets.}
    \label{table_denoising}
    \renewcommand{\arraystretch}{1.5}
    \resizebox{\linewidth}{!}{
    \begin{tabular}{ccccccc}
    \toprule   
    \multirow{2}{*}{Method} & \multicolumn{3}{c}{BSD68} & \multicolumn{3}{c}{Urban100} \\
    \cmidrule(lr){2-4} \cmidrule(lr){5-7} 
     & $\sigma = 15$ & $\sigma = 25$ & $\sigma = 50$ & $\sigma = 15$ & $\sigma = 25$ & $\sigma = 50$\\
    \hline
    CBM3D \cite{cbm3d} & 33.50/0.922 & 30.69/0.868 & 27.36/0.763 & 33.93/0.941 & 31.36/0.909 & 27.93/0.840 \\
    DnCNN \cite{dncnn} & 33.89/0.930 & 31.23/0.883 & 27.92/0.789 & 32.98/0.931 & 30.81/0.902 & 27.59/0.833 \\
    IRCNN \cite{ircnn} & 33.87/0.929 & 31.18/0.882 & 27.88/0.790 & 27.59/0.833 & 31.20/0.909 & 27.70/0.840 \\
    FFDNet \cite{ffdnet} & 33.87/0.929 & 31.21/0.882 & 27.96/0.789 & 33.83/0.942 & 31.40/0.912 & 28.05/0.848 \\
    BRDNet \cite{brdnet} & 34.10/0.929 & 31.43/0.885 & 28.16/0.794 & 34.42/0.946 & 31.99/0.919 & 28.56/0.858 \\
    Restormer \cite{restormer} & 34.29/0.937 & 31.64/0.895 & 28.41/0.810  & 34.67/ \textbf{0.969} & 32.41/0.927 & 29.31/0.878\\
    TKMANet \cite{TKMANet} & - &  - &  - &  33.84/0.941 & 30.34/0.860 & 26.86/0.780 \\
    AirNet \cite{airnet} & 34.14/0.936 & 31.48/0.893 & 28.23/0.806 & 34.40/0.949 & 32.10/0.924 & 28.88/0.871 \\
    DA-CLIP \cite{daclip} & 26.34/0.682 & 25.77/0.653 & 24.31/0.571 & - &  - &  - \\
    PromptIR \cite{promptir} & \textbf{34.34}/\underline{0.938} & \underline{31.71}/\underline{0.897} & \textbf{28.49}/\underline{0.813} & \underline{34.77}/0.952 & \underline{32.49}/\underline{0.929} & \underline{29.39}/\underline{0.881}\\
    \textbf{DINO-IR (Ours)} & \textbf{34.34}/\textbf{0.943} & \textbf{31.72}/\textbf{0.903} & \underline{28.46}/\textbf{0.821} & \textbf{34.91}/\underline{0.961} & \textbf{32.68}/\textbf{0.941} & \textbf{29.60}/\textbf{0.896} \\
    \hline
    \end{tabular}}
\end{table*}

\begin{figure*}[t]
\centering
\includegraphics[width=1\linewidth]{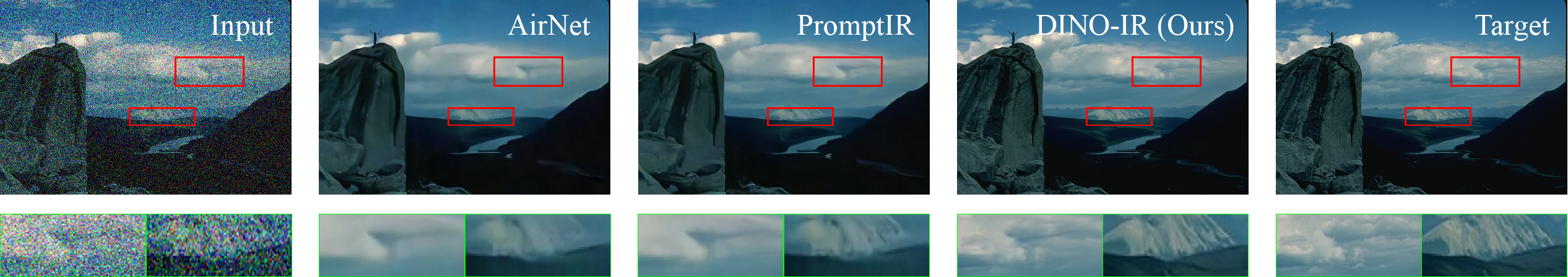}
\caption{\label{image_denoising} Dehazing results on SOTS \cite{sots}. The result images of our method are closer to the target images, achieving a more thorough defogging effect.}
\end{figure*}

\begin{figure*}[t]
\centering
\includegraphics[width=1\linewidth]{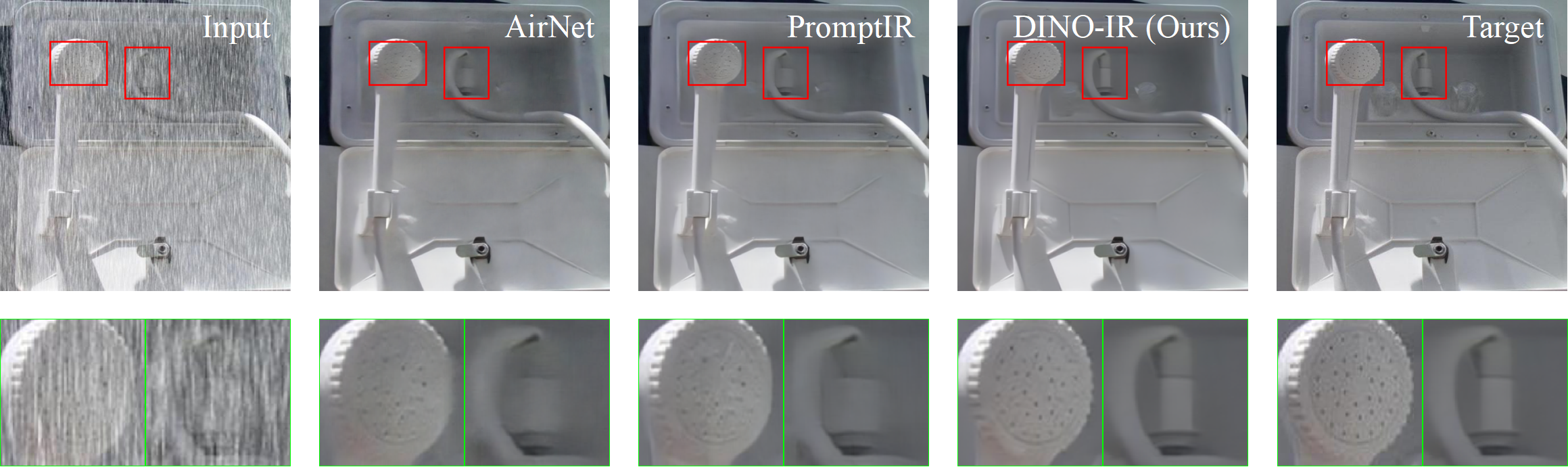}
\caption{\label{image_deraining} Visual under multi-task setting for deraining \cite{did} task. Our method has significant advantages in restoring background rain patterns.}
\end{figure*}

We first describe the utilized datasets and present the implementation details. Subsequently, we offer qualitative and quantitative comparison results with state-of-the-art methods. Finally, we validate the potential and effectiveness of DINOv2 through several ablation experiments.

\subsection{Settings}

\textbf{Datasets and Metrics.} We train our method on a combined dataset comprising four types of degradation: noisy, blurry, rainy, and hazy. Following the previous work \cite{Park}, we use the DID and OTS training datasets \cite{did, sots} to ensure a balanced number of datasets for different tasks. For deraining, we incorporate various levels of rain from the DID \cite{did} dataset for training and testing. In the case of denoising, we generate Gaussian noisy images from the DID \cite{did} with standard deviations ranging from 15 to 50, testing on the BSD68 dataset \cite{bsd68} with the same noise type. For debluring, we produce blurry images from the DID \cite{did} with the Gaussian kernel of size $k = 15$ and random $\sigma\in$ [2.0, 3.1]. For dehazing, we select 4,000 images from the OTS dataset \cite{sots} for training and testing on SOTS \cite{sots}. 

To comprehensively validate DINO-IR, we evaluate some single tasks. As the limited space, we only show the denoising task, and tasks are presented in the supplementary material.
%
%
For single-task image denoising, we use a combination of the BSD400 \cite{bsd400} and WED \cite{wed} datasets for training. The BSD400 \cite{bsd400} dataset includes 400 training images, while the WED \cite{wed} dataset contains 4,744 images. We introduce Gaussian noise at levels denoted by $\sigma$ (from the set {15, 25, 50}) to these clean images for generating training data. 
Performance evaluations are carried out on the BSD68 \cite{bsd68} and Urban100 \cite{urban100} datasets. 
%

{\flushleft \textbf{Implementation Details.}} DINO-IR is based on Restormer \cite{restormer}, it comprises a 4-level encoder-decoder, each level: [4, 6, 6, 8] from level-1 to level-4. The framework is implemented with the Pytorch and trained on two GeForce RTX 3090 GPUs. The batch size is 12, and the patch size is 128. 
We employ the Adam optimizer with $\beta_{1}$ = 0.9, $\beta_{2}$ = 0.999 optimize the model. The initial learning rate is set at $2 \times 10^{-4}$ and reduced to $1 \times 10^{-4}$, adjusting the batch size to 8 and the patch size to 160 for multi-task image restoration. 
The batch size is 4, and the patch size is 128 with a single GeForce RTX 3090 for single-task image restoration. The weight parameter $\lambda$ in~\eqref{L} is empirically set to be $1$. 
The detailed structures of the proposed modules are provided in the supplementary material.

{\flushleft \textbf{Evaluated Methods.}} 
We evaluate methods that provide codes and pre-trained models to ensure fair and comprehensive comparisons. 
For multi-task image restoration, we evaluate both classic and latest approaches, including DnCNN \cite{dncnn}, MPRNet \cite{mprnet}, Restormer \cite{restormer}, NAFNet \cite{nafnet}, AirNet \cite{airnet}, WGWS \cite{wgws}, PromptIR \cite{promptir} and DA-CLIP \cite{daclip}.
It is worth mentioning that the DA-CLIP \cite{daclip} also uses pre-trained models for image restoration. 
However, it focuses on learning degradation categories by textual information from CLIP \cite{clip} while we are exploring the impact of robust features of the DINOv2 \cite{dinov2} on image restoration. 
In the single-task setting, we integrate specific models designed for particular tasks, such as denoising \cite{cbm3d, dncnn, ircnn, ffdnet, brdnet, TKMANet} methods, 
in addition to the aforementioned approaches.

\subsection{Multi-task Image Restoration Results}

%
Table \ref{multi} shows that DINO-IR achieves the best results across multiple test sets. Specifically, DINO-IR yields improved performance by 0.13 dB, 0.96 dB, 0.71 dB, and 1 dB on average of PSNR in four tasks compared to the PromptIR \cite{promptir}, DA-CLIP \cite{daclip}, WGWS \cite{wgws} and AirNet \cite{airnet}, respectively. Furthermore, DINO-IR gains 0.34, 0.88, and 3.41 dB compared to PromptIR \cite{promptir}, DA-CLIP \cite{daclip}, and AirNet \cite{airnet} on the dehazing task. For deraining, our DINO-IR provides 1.14 and 0.75 dB boost over the PromptIR \cite{promptir} and AirNet \cite{airnet}. From the visual results shown in Figs. \ref{image_deraining} and \ref{image_denoising}, DINO-IR is visually favorable against previous state-of-the-art AirNet \cite{airnet} and PromptIR \cite{promptir}.

\subsection{Single-task Image Restoration Results}

We evaluate the performance of our DINO-IR under the single-task setting; a separate model is trained for different restoration tasks. Some experimental results are from PromptIR \cite{promptir}. Table \ref{table_denoising} shows that DINO-IR is comparable with PromptIR \cite{promptir} on BSD68 \cite{bsd68}. However, on Urban100 \cite{urban100}, our method provides respective PSNR gains of 0.14 dB, 0.19 dB, and 0.21 dB over PromptIR \cite{promptir}. This also highlights the generalizability of DINO-IR on other test datasets. Other results are provided in the supplementary material. 

\subsection{Ablation Studies}

{\flushleft \textbf{Effectiveness of the proposed PSF Module and DPC-Loss.}}  In Table \ref{xiaorong}, we present the effects of the PSF module and DPC-Loss on multi-task image restoration. After adding the PSF module, the PSNR is improved by 0.23/0.14, 0.27, and 0.12 dB for deraining, deblurring, and dehazing tasks, respectively, verifying the effectiveness of the PSF module. Table \ref{xiaorong} shows that the DPC-Loss in our framework provides a favorable gain of 0.11/0.07 dB, 0.15 dB, and 0.07 dB for deraining, deblurring, and dehazing tasks.

\begin{table*}[ht]
    \fontsize{16}{16}\selectfont 
    \caption{\label{xiaorong} Effectiveness of PSF module and DPC-Loss.}
    \renewcommand{\arraystretch}{1}
    \resizebox{\linewidth}{!}{
    \begin{tabular}{cccccccc}
        \toprule
        \multirow{2}{*}{Method} & \multicolumn{2}{c}{Deraining} & \multicolumn{3}{c}{Denoising} & \multirow{2}{*}{Deblurring} & \multirow{2}{*}{Dehazing} \\
        \cmidrule(lr){2-3} \cmidrule(lr){4-6} 
         & High & Medium & $\sigma = 15$ & $\sigma = 25$ & $\sigma = 50$ & \\
        \midrule
        Baseline & 30.10/0.887 & 32.15/0.920 & 33.78/0.936 & 31.18/0.892 & 27.96/0.800 & 31.71/0.899 & 29.67/0.975 \\
        with PSF & 30.33/0.891 & 32.29/0.921 & 33.82/0.937 & 31.23/0.893 & 28.01/0.803 & 31.98/0.902 & 29.79/0.975 \\
        with DPC & 30.21/0.888 & 32.22/0.920 & 33.81/0.936 & 31.21/0.893 & 28.00/0.803 & 31.86/0.900 & 29.75/0.975 \\
        \textbf{DINO-IR (Ours)} & \textbf{30.44/0.892}& \textbf{32.37/0.923}& \textbf{33.85/0.938} & \textbf{31.26/0.895} & \textbf{28.04/0.807} & \textbf{32.11/0.903}& \textbf{29.86/0.976} \\
        \bottomrule
    \end{tabular}}
\end{table*}

\begin{table}[ht]
    \fontsize{8}{8}\selectfont 
    \caption{\label{fanhua} Deraining comparisons on unseen dataset Rain100 \cite{rain100} and denoising comparisons on unseen noise level of $\sigma$ = 100 on BSD68 \cite{bsd68}.}
    \renewcommand{\arraystretch}{1.5}
    \resizebox{\linewidth}{!}{
\begin{tabular}{ccc}
\toprule
Methods & $Rain100$ & $\sigma = 100$  \\
\midrule
AIRNet \cite{airnet} & 23.22/0.702 & 23.71/0.683 \\
PromptIR \cite{promptir} & 23.65/0.753& 23.94/0.701\\
\textbf{DINO-IR (Ours)} & \textbf{24.24}/\textbf{0.810}& \textbf{24.06}/\textbf{0.712}\\
        \hline
    \end{tabular}}
\end{table}

\begin{table}[ht]
    \fontsize{8}{8}\selectfont 
    \caption{\label{layer} Effectiveness of different layers from DINOv2. Results are reported on Rain100L.}
    \renewcommand{\arraystretch}{1.5}
    \resizebox{\linewidth}{!}{
    \begin{tabular}{c|cccc}
        Methods & Shallow &  Medium &  Deep &  PSNR(dB) \\
        \hline
        V0 & & & & 36.74\\
        V1 & \checkmark & & & 36.93 \\
        V2 & & \checkmark & & 36.98 \\
        V3 & & & \checkmark & 37.09 \\
        \hline
    \end{tabular}}
\end{table}

\begin{table}[ht]
    \fontsize{6}{6}\selectfont 
    \caption{\label{duo} Effectiveness of PSF module and DPC-Loss.}
    \renewcommand{\arraystretch}{1.5}
    \resizebox{\linewidth}{!}{
    \begin{tabular}{c|cc}
         Methods &  Baseline &  DINO-IR \\
        \hline
        1 (B) & 32.08/0.902 &  32.24/0.904 \\
        2 (B + N) & 31.99/0.901 & 32.18/0.904\\
        3 (B + N + R) & 31.84/0.901 & 32.13/0.903\\
        4 (B + N + R + H) & 31.71/0.899 & 32.11/0.903\\
        \hline
    \end{tabular}}
\end{table}

{\flushleft \textbf{Analysis on Generalizability.}} To evaluate the generalizability of our DINO-IR, we train the model on combined datasets with four types of degradation: 1. testing on unseen dataset Rain100L \cite{rain100} for deraining; 2. testing on unseen degradation level $\sigma$ = 100 for denoising. Table \ref{fanhua} shows that our DINO-IR gets better results on unseen datasets, compared to AirNet \cite{airnet} and PromptIR \cite{promptir}, yielding 1.02/0.35 and 0.59/0.12 dB PSNR gains on both two settings.

{\flushleft \textbf{Analysis on Different Layers of DINOv2.}} In Table \ref{layer}, we present the effects of various DINOv2 layers on the deraining task. The baseline (Restormer \cite{restormer}) for restoration is established as V0. Versions V1, V2, and V3 represent restoration models guided by shallow, medium, and deep layers of DINOv2, respectively. The comparison of V0 with V1, V2, and V3 shows that DINOv2 features enhance restoration with increasing effectiveness in deeper layers. In our framework, we output the scores from the Gating Net in PSF: 0.30, 0.32, and 0.38, which verifies that all three features have important roles and PSF can adaptively learn the corresponding weights to enhance the performance of image restoration.

{\flushleft \textbf{Training model with different tasks.}} In this section, we validate the robustness of DINO-IR (using deblurring as an example). As shown in Table \ref{duo},  B represents deblurring, N denotes denoising, R for deraining, and H signifies dehazing. As the number of tasks increases, DINO-IR exhibits less performance degradation than the baseline, emphasizing its overall robust characteristics.


\section{Conclusion}
\label{Conclusion}

Based on the observation that the features of DINOv2 are independent of degradation factors, we have presented an effective multi-task image restoration approach. The proposed approach, i.e., \mbox{\textbf{DINO-IR}} explores DINOv2 features by the Pixel-semantic fusion, DINO-Restore adaption and fusion module, and DPC loss. Extensive experiments show that our method achieves favorable performance on multi-task image restoration. Moreover, we show that the proposed method performs better results on unseen datasets.

\bibliography{aaai25}

\end{document}